\documentclass[letterpaper]{article} 
\usepackage[submission]{aaai23}  
\usepackage{times}  
\usepackage{helvet}  
\usepackage{courier}  
\usepackage[hyphens]{url}  
\usepackage{graphicx} 
\usepackage{natbib}  
\usepackage{caption} 
\usepackage{algorithm}
\usepackage{algorithmicx}
\usepackage[noend]{algpseudocode}

\usepackage{bm}

\usepackage{newfloat}
\usepackage{listings}
\DeclareCaptionStyle{ruled}{labelfont=normalfont,labelsep=colon,strut=off} 
\floatstyle{ruled}
\newfloat{listing}{tb}{lst}{}
\floatname{listing}{Listing}
\usepackage{amsfonts}

\usepackage{mathtools}
\DeclarePairedDelimiterX{\infdivx}[2]{(}{)}{%
  #1\;\delimsize\|\;#2%
}

\usepackage{cleveref}
\usepackage{dsfont}
\usepackage{siunitx}
  
  \DeclareMathOperator*{\argmax}{arg\,max}
\DeclareMathOperator*{\argmin}{arg\,min}

\newcommand{\I}[1]{\mathds{1}_{\{#1\}}}

\usepackage{amsthm}
\usepackage{booktabs}

\usepackage{tikz}
\usepackage{pgfplots}
\pgfplotsset{compat=1.17}
\usetikzlibrary{arrows.meta}
\usetikzlibrary{backgrounds}
\usepgfplotslibrary{patchplots}
\usepgfplotslibrary{fillbetween}
\pgfplotsset{%
layers/standard/.define layer set={%
    background,axis background,axis grid,axis ticks,axis lines,axis tick labels,pre main,main,axis descriptions,axis foreground%
}{grid style= {/pgfplots/on layer=axis grid},%
    tick style= {/pgfplots/on layer=axis ticks},%
    axis line style= {/pgfplots/on layer=axis lines},%
    label style= {/pgfplots/on layer=axis descriptions},%
    legend style= {/pgfplots/on layer=axis descriptions},%
    title style= {/pgfplots/on layer=axis descriptions},%
    colorbar style= {/pgfplots/on layer=axis descriptions},%
    ticklabel style= {/pgfplots/on layer=axis tick labels},%
    axis background@ style={/pgfplots/on layer=axis background},%
    3d box foreground style={/pgfplots/on layer=axis foreground},%
    },
}
\pgfplotsset{every axis legend/.append style={legend cell align=left}}

\usetikzlibrary{arrows.meta}
\usetikzlibrary{backgrounds}
\usepgfplotslibrary{patchplots}
\usepgfplotslibrary{fillbetween}
\pgfplotsset{%
    layers/standard/.define layer set={%
        background,axis background,axis grid,axis ticks,axis lines,axis tick labels,pre main,main,axis descriptions,axis foreground%
    }{
        grid style={/pgfplots/on layer=axis grid},%
        tick style={/pgfplots/on layer=axis ticks},%
        axis line style={/pgfplots/on layer=axis lines},%
        label style={/pgfplots/on layer=axis descriptions},%
        legend style={/pgfplots/on layer=axis descriptions},%
        title style={/pgfplots/on layer=axis descriptions},%
        colorbar style={/pgfplots/on layer=axis descriptions},%
        ticklabel style={/pgfplots/on layer=axis tick labels},%
        axis background@ style={/pgfplots/on layer=axis background},%
        3d box foreground style={/pgfplots/on layer=axis foreground},%
    },
}

\pgfplotsset{
colormap={plots1}{rgb(0.00000000)=(0.26700400,0.00487400,0.32941500)
rgb(0.00392157)=(0.26851000,0.00960500,0.33542700)
rgb(0.00784314)=(0.26994400,0.01462500,0.34137900)
rgb(0.01176471)=(0.27130500,0.01994200,0.34726900)
rgb(0.01568627)=(0.27259400,0.02556300,0.35309300)
rgb(0.01960784)=(0.27380900,0.03149700,0.35885300)
rgb(0.02352941)=(0.27495200,0.03775200,0.36454300)
rgb(0.02745098)=(0.27602200,0.04416700,0.37016400)
rgb(0.03137255)=(0.27701800,0.05034400,0.37571500)
rgb(0.03529412)=(0.27794100,0.05632400,0.38119100)
rgb(0.03921569)=(0.27879100,0.06214500,0.38659200)
rgb(0.04313725)=(0.27956600,0.06783600,0.39191700)
rgb(0.04705882)=(0.28026700,0.07341700,0.39716300)
rgb(0.05098039)=(0.28089400,0.07890700,0.40232900)
rgb(0.05490196)=(0.28144600,0.08432000,0.40741400)
rgb(0.05882353)=(0.28192400,0.08966600,0.41241500)
rgb(0.06274510)=(0.28232700,0.09495500,0.41733100)
rgb(0.06666667)=(0.28265600,0.10019600,0.42216000)
rgb(0.07058824)=(0.28291000,0.10539300,0.42690200)
rgb(0.07450980)=(0.28309100,0.11055300,0.43155400)
rgb(0.07843137)=(0.28319700,0.11568000,0.43611500)
rgb(0.08235294)=(0.28322900,0.12077700,0.44058400)
rgb(0.08627451)=(0.28318700,0.12584800,0.44496000)
rgb(0.09019608)=(0.28307200,0.13089500,0.44924100)
rgb(0.09411765)=(0.28288400,0.13592000,0.45342700)
rgb(0.09803922)=(0.28262300,0.14092600,0.45751700)
rgb(0.10196078)=(0.28229000,0.14591200,0.46151000)
rgb(0.10588235)=(0.28188700,0.15088100,0.46540500)
rgb(0.10980392)=(0.28141200,0.15583400,0.46920100)
rgb(0.11372549)=(0.28086800,0.16077100,0.47289900)
rgb(0.11764706)=(0.28025500,0.16569300,0.47649800)
rgb(0.12156863)=(0.27957400,0.17059900,0.47999700)
rgb(0.12549020)=(0.27882600,0.17549000,0.48339700)
rgb(0.12941176)=(0.27801200,0.18036700,0.48669700)
rgb(0.13333333)=(0.27713400,0.18522800,0.48989800)
rgb(0.13725490)=(0.27619400,0.19007400,0.49300100)
rgb(0.14117647)=(0.27519100,0.19490500,0.49600500)
rgb(0.14509804)=(0.27412800,0.19972100,0.49891100)
rgb(0.14901961)=(0.27300600,0.20452000,0.50172100)
rgb(0.15294118)=(0.27182800,0.20930300,0.50443400)
rgb(0.15686275)=(0.27059500,0.21406900,0.50705200)
rgb(0.16078431)=(0.26930800,0.21881800,0.50957700)
rgb(0.16470588)=(0.26796800,0.22354900,0.51200800)
rgb(0.16862745)=(0.26658000,0.22826200,0.51434900)
rgb(0.17254902)=(0.26514500,0.23295600,0.51659900)
rgb(0.17647059)=(0.26366300,0.23763100,0.51876200)
rgb(0.18039216)=(0.26213800,0.24228600,0.52083700)
rgb(0.18431373)=(0.26057100,0.24692200,0.52282800)
rgb(0.18823529)=(0.25896500,0.25153700,0.52473600)
rgb(0.19215686)=(0.25732200,0.25613000,0.52656300)
rgb(0.19607843)=(0.25564500,0.26070300,0.52831200)
rgb(0.20000000)=(0.25393500,0.26525400,0.52998300)
rgb(0.20392157)=(0.25219400,0.26978300,0.53157900)
rgb(0.20784314)=(0.25042500,0.27429000,0.53310300)
rgb(0.21176471)=(0.24862900,0.27877500,0.53455600)
rgb(0.21568627)=(0.24681100,0.28323700,0.53594100)
rgb(0.21960784)=(0.24497200,0.28767500,0.53726000)
rgb(0.22352941)=(0.24311300,0.29209200,0.53851600)
rgb(0.22745098)=(0.24123700,0.29648500,0.53970900)
rgb(0.23137255)=(0.23934600,0.30085500,0.54084400)
rgb(0.23529412)=(0.23744100,0.30520200,0.54192100)
rgb(0.23921569)=(0.23552600,0.30952700,0.54294400)
rgb(0.24313725)=(0.23360300,0.31382800,0.54391400)
rgb(0.24705882)=(0.23167400,0.31810600,0.54483400)
rgb(0.25098039)=(0.22973900,0.32236100,0.54570600)
rgb(0.25490196)=(0.22780200,0.32659400,0.54653200)
rgb(0.25882353)=(0.22586300,0.33080500,0.54731400)
rgb(0.26274510)=(0.22392500,0.33499400,0.54805300)
rgb(0.26666667)=(0.22198900,0.33916100,0.54875200)
rgb(0.27058824)=(0.22005700,0.34330700,0.54941300)
rgb(0.27450980)=(0.21813000,0.34743200,0.55003800)
rgb(0.27843137)=(0.21621000,0.35153500,0.55062700)
rgb(0.28235294)=(0.21429800,0.35561900,0.55118400)
rgb(0.28627451)=(0.21239500,0.35968300,0.55171000)
rgb(0.29019608)=(0.21050300,0.36372700,0.55220600)
rgb(0.29411765)=(0.20862300,0.36775200,0.55267500)
rgb(0.29803922)=(0.20675600,0.37175800,0.55311700)
rgb(0.30196078)=(0.20490300,0.37574600,0.55353300)
rgb(0.30588235)=(0.20306300,0.37971600,0.55392500)
rgb(0.30980392)=(0.20123900,0.38367000,0.55429400)
rgb(0.31372549)=(0.19943000,0.38760700,0.55464200)
rgb(0.31764706)=(0.19763600,0.39152800,0.55496900)
rgb(0.32156863)=(0.19586000,0.39543300,0.55527600)
rgb(0.32549020)=(0.19410000,0.39932300,0.55556500)
rgb(0.32941176)=(0.19235700,0.40319900,0.55583600)
rgb(0.33333333)=(0.19063100,0.40706100,0.55608900)
rgb(0.33725490)=(0.18892300,0.41091000,0.55632600)
rgb(0.34117647)=(0.18723100,0.41474600,0.55654700)
rgb(0.34509804)=(0.18555600,0.41857000,0.55675300)
rgb(0.34901961)=(0.18389800,0.42238300,0.55694400)
rgb(0.35294118)=(0.18225600,0.42618400,0.55712000)
rgb(0.35686275)=(0.18062900,0.42997500,0.55728200)
rgb(0.36078431)=(0.17901900,0.43375600,0.55743000)
rgb(0.36470588)=(0.17742300,0.43752700,0.55756500)
rgb(0.36862745)=(0.17584100,0.44129000,0.55768500)
rgb(0.37254902)=(0.17427400,0.44504400,0.55779200)
rgb(0.37647059)=(0.17271900,0.44879100,0.55788500)
rgb(0.38039216)=(0.17117600,0.45253000,0.55796500)
rgb(0.38431373)=(0.16964600,0.45626200,0.55803000)
rgb(0.38823529)=(0.16812600,0.45998800,0.55808200)
rgb(0.39215686)=(0.16661700,0.46370800,0.55811900)
rgb(0.39607843)=(0.16511700,0.46742300,0.55814100)
rgb(0.40000000)=(0.16362500,0.47113300,0.55814800)
rgb(0.40392157)=(0.16214200,0.47483800,0.55814000)
rgb(0.40784314)=(0.16066500,0.47854000,0.55811500)
rgb(0.41176471)=(0.15919400,0.48223700,0.55807300)
rgb(0.41568627)=(0.15772900,0.48593200,0.55801300)
rgb(0.41960784)=(0.15627000,0.48962400,0.55793600)
rgb(0.42352941)=(0.15481500,0.49331300,0.55784000)
rgb(0.42745098)=(0.15336400,0.49700000,0.55772400)
rgb(0.43137255)=(0.15191800,0.50068500,0.55758700)
rgb(0.43529412)=(0.15047600,0.50436900,0.55743000)
rgb(0.43921569)=(0.14903900,0.50805100,0.55725000)
rgb(0.44313725)=(0.14760700,0.51173300,0.55704900)
rgb(0.44705882)=(0.14618000,0.51541300,0.55682300)
rgb(0.45098039)=(0.14475900,0.51909300,0.55657200)
rgb(0.45490196)=(0.14334300,0.52277300,0.55629500)
rgb(0.45882353)=(0.14193500,0.52645300,0.55599100)
rgb(0.46274510)=(0.14053600,0.53013200,0.55565900)
rgb(0.46666667)=(0.13914700,0.53381200,0.55529800)
rgb(0.47058824)=(0.13777000,0.53749200,0.55490600)
rgb(0.47450980)=(0.13640800,0.54117300,0.55448300)
rgb(0.47843137)=(0.13506600,0.54485300,0.55402900)
rgb(0.48235294)=(0.13374300,0.54853500,0.55354100)
rgb(0.48627451)=(0.13244400,0.55221600,0.55301800)
rgb(0.49019608)=(0.13117200,0.55589900,0.55245900)
rgb(0.49411765)=(0.12993300,0.55958200,0.55186400)
rgb(0.49803922)=(0.12872900,0.56326500,0.55122900)
rgb(0.50196078)=(0.12756800,0.56694900,0.55055600)
rgb(0.50588235)=(0.12645300,0.57063300,0.54984100)
rgb(0.50980392)=(0.12539400,0.57431800,0.54908600)
rgb(0.51372549)=(0.12439500,0.57800200,0.54828700)
rgb(0.51764706)=(0.12346300,0.58168700,0.54744500)
rgb(0.52156863)=(0.12260600,0.58537100,0.54655700)
rgb(0.52549020)=(0.12183100,0.58905500,0.54562300)
rgb(0.52941176)=(0.12114800,0.59273900,0.54464100)
rgb(0.53333333)=(0.12056500,0.59642200,0.54361100)
rgb(0.53725490)=(0.12009200,0.60010400,0.54253000)
rgb(0.54117647)=(0.11973800,0.60378500,0.54140000)
rgb(0.54509804)=(0.11951200,0.60746400,0.54021800)
rgb(0.54901961)=(0.11942300,0.61114100,0.53898200)
rgb(0.55294118)=(0.11948300,0.61481700,0.53769200)
rgb(0.55686275)=(0.11969900,0.61849000,0.53634700)
rgb(0.56078431)=(0.12008100,0.62216100,0.53494600)
rgb(0.56470588)=(0.12063800,0.62582800,0.53348800)
rgb(0.56862745)=(0.12138000,0.62949200,0.53197300)
rgb(0.57254902)=(0.12231200,0.63315300,0.53039800)
rgb(0.57647059)=(0.12344400,0.63680900,0.52876300)
rgb(0.58039216)=(0.12478000,0.64046100,0.52706800)
rgb(0.58431373)=(0.12632600,0.64410700,0.52531100)
rgb(0.58823529)=(0.12808700,0.64774900,0.52349100)
rgb(0.59215686)=(0.13006700,0.65138400,0.52160800)
rgb(0.59607843)=(0.13226800,0.65501400,0.51966100)
rgb(0.60000000)=(0.13469200,0.65863600,0.51764900)
rgb(0.60392157)=(0.13733900,0.66225200,0.51557100)
rgb(0.60784314)=(0.14021000,0.66585900,0.51342700)
rgb(0.61176471)=(0.14330300,0.66945900,0.51121500)
rgb(0.61568627)=(0.14661600,0.67305000,0.50893600)
rgb(0.61960784)=(0.15014800,0.67663100,0.50658900)
rgb(0.62352941)=(0.15389400,0.68020300,0.50417200)
rgb(0.62745098)=(0.15785100,0.68376500,0.50168600)
rgb(0.63137255)=(0.16201600,0.68731600,0.49912900)
rgb(0.63529412)=(0.16638300,0.69085600,0.49650200)
rgb(0.63921569)=(0.17094800,0.69438400,0.49380300)
rgb(0.64313725)=(0.17570700,0.69790000,0.49103300)
rgb(0.64705882)=(0.18065300,0.70140200,0.48818900)
rgb(0.65098039)=(0.18578300,0.70489100,0.48527300)
rgb(0.65490196)=(0.19109000,0.70836600,0.48228400)
rgb(0.65882353)=(0.19657100,0.71182700,0.47922100)
rgb(0.66274510)=(0.20221900,0.71527200,0.47608400)
rgb(0.66666667)=(0.20803000,0.71870100,0.47287300)
rgb(0.67058824)=(0.21400000,0.72211400,0.46958800)
rgb(0.67450980)=(0.22012400,0.72550900,0.46622600)
rgb(0.67843137)=(0.22639700,0.72888800,0.46278900)
rgb(0.68235294)=(0.23281500,0.73224700,0.45927700)
rgb(0.68627451)=(0.23937400,0.73558800,0.45568800)
rgb(0.69019608)=(0.24607000,0.73891000,0.45202400)
rgb(0.69411765)=(0.25289900,0.74221100,0.44828400)
rgb(0.69803922)=(0.25985700,0.74549200,0.44446700)
rgb(0.70196078)=(0.26694100,0.74875100,0.44057300)
rgb(0.70588235)=(0.27414900,0.75198800,0.43660100)
rgb(0.70980392)=(0.28147700,0.75520300,0.43255200)
rgb(0.71372549)=(0.28892100,0.75839400,0.42842600)
rgb(0.71764706)=(0.29647900,0.76156100,0.42422300)
rgb(0.72156863)=(0.30414800,0.76470400,0.41994300)
rgb(0.72549020)=(0.31192500,0.76782200,0.41558600)
rgb(0.72941176)=(0.31980900,0.77091400,0.41115200)
rgb(0.73333333)=(0.32779600,0.77398000,0.40664000)
rgb(0.73725490)=(0.33588500,0.77701800,0.40204900)
rgb(0.74117647)=(0.34407400,0.78002900,0.39738100)
rgb(0.74509804)=(0.35236000,0.78301100,0.39263600)
rgb(0.74901961)=(0.36074100,0.78596400,0.38781400)
rgb(0.75294118)=(0.36921400,0.78888800,0.38291400)
rgb(0.75686275)=(0.37777900,0.79178100,0.37793900)
rgb(0.76078431)=(0.38643300,0.79464400,0.37288600)
rgb(0.76470588)=(0.39517400,0.79747500,0.36775700)
rgb(0.76862745)=(0.40400100,0.80027500,0.36255200)
rgb(0.77254902)=(0.41291300,0.80304100,0.35726900)
rgb(0.77647059)=(0.42190800,0.80577400,0.35191000)
rgb(0.78039216)=(0.43098300,0.80847300,0.34647600)
rgb(0.78431373)=(0.44013700,0.81113800,0.34096700)
rgb(0.78823529)=(0.44936800,0.81376800,0.33538400)
rgb(0.79215686)=(0.45867400,0.81636300,0.32972700)
rgb(0.79607843)=(0.46805300,0.81892100,0.32399800)
rgb(0.80000000)=(0.47750400,0.82144400,0.31819500)
rgb(0.80392157)=(0.48702600,0.82392900,0.31232100)
rgb(0.80784314)=(0.49661500,0.82637600,0.30637700)
rgb(0.81176471)=(0.50627100,0.82878600,0.30036200)
rgb(0.81568627)=(0.51599200,0.83115800,0.29427900)
rgb(0.81960784)=(0.52577600,0.83349100,0.28812700)
rgb(0.82352941)=(0.53562100,0.83578500,0.28190800)
rgb(0.82745098)=(0.54552400,0.83803900,0.27562600)
rgb(0.83137255)=(0.55548400,0.84025400,0.26928100)
rgb(0.83529412)=(0.56549800,0.84243000,0.26287700)
rgb(0.83921569)=(0.57556300,0.84456600,0.25641500)
rgb(0.84313725)=(0.58567800,0.84666100,0.24989700)
rgb(0.84705882)=(0.59583900,0.84871700,0.24332900)
rgb(0.85098039)=(0.60604500,0.85073300,0.23671200)
rgb(0.85490196)=(0.61629300,0.85270900,0.23005200)
rgb(0.85882353)=(0.62657900,0.85464500,0.22335300)
rgb(0.86274510)=(0.63690200,0.85654200,0.21662000)
rgb(0.86666667)=(0.64725700,0.85840000,0.20986100)
rgb(0.87058824)=(0.65764200,0.86021900,0.20308200)
rgb(0.87450980)=(0.66805400,0.86199900,0.19629300)
rgb(0.87843137)=(0.67848900,0.86374200,0.18950300)
rgb(0.88235294)=(0.68894400,0.86544800,0.18272500)
rgb(0.88627451)=(0.69941500,0.86711700,0.17597100)
rgb(0.89019608)=(0.70989800,0.86875100,0.16925700)
rgb(0.89411765)=(0.72039100,0.87035000,0.16260300)
rgb(0.89803922)=(0.73088900,0.87191600,0.15602900)
rgb(0.90196078)=(0.74138800,0.87344900,0.14956100)
rgb(0.90588235)=(0.75188400,0.87495100,0.14322800)
rgb(0.90980392)=(0.76237300,0.87642400,0.13706400)
rgb(0.91372549)=(0.77285200,0.87786800,0.13110900)
rgb(0.91764706)=(0.78331500,0.87928500,0.12540500)
rgb(0.92156863)=(0.79376000,0.88067800,0.12000500)
rgb(0.92549020)=(0.80418200,0.88204600,0.11496500)
rgb(0.92941176)=(0.81457600,0.88339300,0.11034700)
rgb(0.93333333)=(0.82494000,0.88472000,0.10621700)
rgb(0.93725490)=(0.83527000,0.88602900,0.10264600)
rgb(0.94117647)=(0.84556100,0.88732200,0.09970200)
rgb(0.94509804)=(0.85581000,0.88860100,0.09745200)
rgb(0.94901961)=(0.86601300,0.88986800,0.09595300)
rgb(0.95294118)=(0.87616800,0.89112500,0.09525000)
rgb(0.95686275)=(0.88627100,0.89237400,0.09537400)
rgb(0.96078431)=(0.89632000,0.89361600,0.09633500)
rgb(0.96470588)=(0.90631100,0.89485500,0.09812500)
rgb(0.96862745)=(0.91624200,0.89609100,0.10071700)
rgb(0.97254902)=(0.92610600,0.89733000,0.10407100)
rgb(0.97647059)=(0.93590400,0.89857000,0.10813100)
rgb(0.98039216)=(0.94563600,0.89981500,0.11283800)
rgb(0.98431373)=(0.95530000,0.90106500,0.11812800)
rgb(0.98823529)=(0.96489400,0.90232300,0.12394100)
rgb(0.99215686)=(0.97441700,0.90359000,0.13021500)
rgb(0.99607843)=(0.98386800,0.90486700,0.13689700)
rgb(1.00000000)=(0.99324800,0.90615700,0.14393600)},
}
\pgfplotsset{
colormap={plots1}{rgb(0.00000000)=(0.26700400,0.00487400,0.32941500)
rgb(0.00392157)=(0.26851000,0.00960500,0.33542700)
rgb(0.00784314)=(0.26994400,0.01462500,0.34137900)
rgb(0.01176471)=(0.27130500,0.01994200,0.34726900)
rgb(0.01568627)=(0.27259400,0.02556300,0.35309300)
rgb(0.01960784)=(0.27380900,0.03149700,0.35885300)
rgb(0.02352941)=(0.27495200,0.03775200,0.36454300)
rgb(0.02745098)=(0.27602200,0.04416700,0.37016400)
rgb(0.03137255)=(0.27701800,0.05034400,0.37571500)
rgb(0.03529412)=(0.27794100,0.05632400,0.38119100)
rgb(0.03921569)=(0.27879100,0.06214500,0.38659200)
rgb(0.04313725)=(0.27956600,0.06783600,0.39191700)
rgb(0.04705882)=(0.28026700,0.07341700,0.39716300)
rgb(0.05098039)=(0.28089400,0.07890700,0.40232900)
rgb(0.05490196)=(0.28144600,0.08432000,0.40741400)
rgb(0.05882353)=(0.28192400,0.08966600,0.41241500)
rgb(0.06274510)=(0.28232700,0.09495500,0.41733100)
rgb(0.06666667)=(0.28265600,0.10019600,0.42216000)
rgb(0.07058824)=(0.28291000,0.10539300,0.42690200)
rgb(0.07450980)=(0.28309100,0.11055300,0.43155400)
rgb(0.07843137)=(0.28319700,0.11568000,0.43611500)
rgb(0.08235294)=(0.28322900,0.12077700,0.44058400)
rgb(0.08627451)=(0.28318700,0.12584800,0.44496000)
rgb(0.09019608)=(0.28307200,0.13089500,0.44924100)
rgb(0.09411765)=(0.28288400,0.13592000,0.45342700)
rgb(0.09803922)=(0.28262300,0.14092600,0.45751700)
rgb(0.10196078)=(0.28229000,0.14591200,0.46151000)
rgb(0.10588235)=(0.28188700,0.15088100,0.46540500)
rgb(0.10980392)=(0.28141200,0.15583400,0.46920100)
rgb(0.11372549)=(0.28086800,0.16077100,0.47289900)
rgb(0.11764706)=(0.28025500,0.16569300,0.47649800)
rgb(0.12156863)=(0.27957400,0.17059900,0.47999700)
rgb(0.12549020)=(0.27882600,0.17549000,0.48339700)
rgb(0.12941176)=(0.27801200,0.18036700,0.48669700)
rgb(0.13333333)=(0.27713400,0.18522800,0.48989800)
rgb(0.13725490)=(0.27619400,0.19007400,0.49300100)
rgb(0.14117647)=(0.27519100,0.19490500,0.49600500)
rgb(0.14509804)=(0.27412800,0.19972100,0.49891100)
rgb(0.14901961)=(0.27300600,0.20452000,0.50172100)
rgb(0.15294118)=(0.27182800,0.20930300,0.50443400)
rgb(0.15686275)=(0.27059500,0.21406900,0.50705200)
rgb(0.16078431)=(0.26930800,0.21881800,0.50957700)
rgb(0.16470588)=(0.26796800,0.22354900,0.51200800)
rgb(0.16862745)=(0.26658000,0.22826200,0.51434900)
rgb(0.17254902)=(0.26514500,0.23295600,0.51659900)
rgb(0.17647059)=(0.26366300,0.23763100,0.51876200)
rgb(0.18039216)=(0.26213800,0.24228600,0.52083700)
rgb(0.18431373)=(0.26057100,0.24692200,0.52282800)
rgb(0.18823529)=(0.25896500,0.25153700,0.52473600)
rgb(0.19215686)=(0.25732200,0.25613000,0.52656300)
rgb(0.19607843)=(0.25564500,0.26070300,0.52831200)
rgb(0.20000000)=(0.25393500,0.26525400,0.52998300)
rgb(0.20392157)=(0.25219400,0.26978300,0.53157900)
rgb(0.20784314)=(0.25042500,0.27429000,0.53310300)
rgb(0.21176471)=(0.24862900,0.27877500,0.53455600)
rgb(0.21568627)=(0.24681100,0.28323700,0.53594100)
rgb(0.21960784)=(0.24497200,0.28767500,0.53726000)
rgb(0.22352941)=(0.24311300,0.29209200,0.53851600)
rgb(0.22745098)=(0.24123700,0.29648500,0.53970900)
rgb(0.23137255)=(0.23934600,0.30085500,0.54084400)
rgb(0.23529412)=(0.23744100,0.30520200,0.54192100)
rgb(0.23921569)=(0.23552600,0.30952700,0.54294400)
rgb(0.24313725)=(0.23360300,0.31382800,0.54391400)
rgb(0.24705882)=(0.23167400,0.31810600,0.54483400)
rgb(0.25098039)=(0.22973900,0.32236100,0.54570600)
rgb(0.25490196)=(0.22780200,0.32659400,0.54653200)
rgb(0.25882353)=(0.22586300,0.33080500,0.54731400)
rgb(0.26274510)=(0.22392500,0.33499400,0.54805300)
rgb(0.26666667)=(0.22198900,0.33916100,0.54875200)
rgb(0.27058824)=(0.22005700,0.34330700,0.54941300)
rgb(0.27450980)=(0.21813000,0.34743200,0.55003800)
rgb(0.27843137)=(0.21621000,0.35153500,0.55062700)
rgb(0.28235294)=(0.21429800,0.35561900,0.55118400)
rgb(0.28627451)=(0.21239500,0.35968300,0.55171000)
rgb(0.29019608)=(0.21050300,0.36372700,0.55220600)
rgb(0.29411765)=(0.20862300,0.36775200,0.55267500)
rgb(0.29803922)=(0.20675600,0.37175800,0.55311700)
rgb(0.30196078)=(0.20490300,0.37574600,0.55353300)
rgb(0.30588235)=(0.20306300,0.37971600,0.55392500)
rgb(0.30980392)=(0.20123900,0.38367000,0.55429400)
rgb(0.31372549)=(0.19943000,0.38760700,0.55464200)
rgb(0.31764706)=(0.19763600,0.39152800,0.55496900)
rgb(0.32156863)=(0.19586000,0.39543300,0.55527600)
rgb(0.32549020)=(0.19410000,0.39932300,0.55556500)
rgb(0.32941176)=(0.19235700,0.40319900,0.55583600)
rgb(0.33333333)=(0.19063100,0.40706100,0.55608900)
rgb(0.33725490)=(0.18892300,0.41091000,0.55632600)
rgb(0.34117647)=(0.18723100,0.41474600,0.55654700)
rgb(0.34509804)=(0.18555600,0.41857000,0.55675300)
rgb(0.34901961)=(0.18389800,0.42238300,0.55694400)
rgb(0.35294118)=(0.18225600,0.42618400,0.55712000)
rgb(0.35686275)=(0.18062900,0.42997500,0.55728200)
rgb(0.36078431)=(0.17901900,0.43375600,0.55743000)
rgb(0.36470588)=(0.17742300,0.43752700,0.55756500)
rgb(0.36862745)=(0.17584100,0.44129000,0.55768500)
rgb(0.37254902)=(0.17427400,0.44504400,0.55779200)
rgb(0.37647059)=(0.17271900,0.44879100,0.55788500)
rgb(0.38039216)=(0.17117600,0.45253000,0.55796500)
rgb(0.38431373)=(0.16964600,0.45626200,0.55803000)
rgb(0.38823529)=(0.16812600,0.45998800,0.55808200)
rgb(0.39215686)=(0.16661700,0.46370800,0.55811900)
rgb(0.39607843)=(0.16511700,0.46742300,0.55814100)
rgb(0.40000000)=(0.16362500,0.47113300,0.55814800)
rgb(0.40392157)=(0.16214200,0.47483800,0.55814000)
rgb(0.40784314)=(0.16066500,0.47854000,0.55811500)
rgb(0.41176471)=(0.15919400,0.48223700,0.55807300)
rgb(0.41568627)=(0.15772900,0.48593200,0.55801300)
rgb(0.41960784)=(0.15627000,0.48962400,0.55793600)
rgb(0.42352941)=(0.15481500,0.49331300,0.55784000)
rgb(0.42745098)=(0.15336400,0.49700000,0.55772400)
rgb(0.43137255)=(0.15191800,0.50068500,0.55758700)
rgb(0.43529412)=(0.15047600,0.50436900,0.55743000)
rgb(0.43921569)=(0.14903900,0.50805100,0.55725000)
rgb(0.44313725)=(0.14760700,0.51173300,0.55704900)
rgb(0.44705882)=(0.14618000,0.51541300,0.55682300)
rgb(0.45098039)=(0.14475900,0.51909300,0.55657200)
rgb(0.45490196)=(0.14334300,0.52277300,0.55629500)
rgb(0.45882353)=(0.14193500,0.52645300,0.55599100)
rgb(0.46274510)=(0.14053600,0.53013200,0.55565900)
rgb(0.46666667)=(0.13914700,0.53381200,0.55529800)
rgb(0.47058824)=(0.13777000,0.53749200,0.55490600)
rgb(0.47450980)=(0.13640800,0.54117300,0.55448300)
rgb(0.47843137)=(0.13506600,0.54485300,0.55402900)
rgb(0.48235294)=(0.13374300,0.54853500,0.55354100)
rgb(0.48627451)=(0.13244400,0.55221600,0.55301800)
rgb(0.49019608)=(0.13117200,0.55589900,0.55245900)
rgb(0.49411765)=(0.12993300,0.55958200,0.55186400)
rgb(0.49803922)=(0.12872900,0.56326500,0.55122900)
rgb(0.50196078)=(0.12756800,0.56694900,0.55055600)
rgb(0.50588235)=(0.12645300,0.57063300,0.54984100)
rgb(0.50980392)=(0.12539400,0.57431800,0.54908600)
rgb(0.51372549)=(0.12439500,0.57800200,0.54828700)
rgb(0.51764706)=(0.12346300,0.58168700,0.54744500)
rgb(0.52156863)=(0.12260600,0.58537100,0.54655700)
rgb(0.52549020)=(0.12183100,0.58905500,0.54562300)
rgb(0.52941176)=(0.12114800,0.59273900,0.54464100)
rgb(0.53333333)=(0.12056500,0.59642200,0.54361100)
rgb(0.53725490)=(0.12009200,0.60010400,0.54253000)
rgb(0.54117647)=(0.11973800,0.60378500,0.54140000)
rgb(0.54509804)=(0.11951200,0.60746400,0.54021800)
rgb(0.54901961)=(0.11942300,0.61114100,0.53898200)
rgb(0.55294118)=(0.11948300,0.61481700,0.53769200)
rgb(0.55686275)=(0.11969900,0.61849000,0.53634700)
rgb(0.56078431)=(0.12008100,0.62216100,0.53494600)
rgb(0.56470588)=(0.12063800,0.62582800,0.53348800)
rgb(0.56862745)=(0.12138000,0.62949200,0.53197300)
rgb(0.57254902)=(0.12231200,0.63315300,0.53039800)
rgb(0.57647059)=(0.12344400,0.63680900,0.52876300)
rgb(0.58039216)=(0.12478000,0.64046100,0.52706800)
rgb(0.58431373)=(0.12632600,0.64410700,0.52531100)
rgb(0.58823529)=(0.12808700,0.64774900,0.52349100)
rgb(0.59215686)=(0.13006700,0.65138400,0.52160800)
rgb(0.59607843)=(0.13226800,0.65501400,0.51966100)
rgb(0.60000000)=(0.13469200,0.65863600,0.51764900)
rgb(0.60392157)=(0.13733900,0.66225200,0.51557100)
rgb(0.60784314)=(0.14021000,0.66585900,0.51342700)
rgb(0.61176471)=(0.14330300,0.66945900,0.51121500)
rgb(0.61568627)=(0.14661600,0.67305000,0.50893600)
rgb(0.61960784)=(0.15014800,0.67663100,0.50658900)
rgb(0.62352941)=(0.15389400,0.68020300,0.50417200)
rgb(0.62745098)=(0.15785100,0.68376500,0.50168600)
rgb(0.63137255)=(0.16201600,0.68731600,0.49912900)
rgb(0.63529412)=(0.16638300,0.69085600,0.49650200)
rgb(0.63921569)=(0.17094800,0.69438400,0.49380300)
rgb(0.64313725)=(0.17570700,0.69790000,0.49103300)
rgb(0.64705882)=(0.18065300,0.70140200,0.48818900)
rgb(0.65098039)=(0.18578300,0.70489100,0.48527300)
rgb(0.65490196)=(0.19109000,0.70836600,0.48228400)
rgb(0.65882353)=(0.19657100,0.71182700,0.47922100)
rgb(0.66274510)=(0.20221900,0.71527200,0.47608400)
rgb(0.66666667)=(0.20803000,0.71870100,0.47287300)
rgb(0.67058824)=(0.21400000,0.72211400,0.46958800)
rgb(0.67450980)=(0.22012400,0.72550900,0.46622600)
rgb(0.67843137)=(0.22639700,0.72888800,0.46278900)
rgb(0.68235294)=(0.23281500,0.73224700,0.45927700)
rgb(0.68627451)=(0.23937400,0.73558800,0.45568800)
rgb(0.69019608)=(0.24607000,0.73891000,0.45202400)
rgb(0.69411765)=(0.25289900,0.74221100,0.44828400)
rgb(0.69803922)=(0.25985700,0.74549200,0.44446700)
rgb(0.70196078)=(0.26694100,0.74875100,0.44057300)
rgb(0.70588235)=(0.27414900,0.75198800,0.43660100)
rgb(0.70980392)=(0.28147700,0.75520300,0.43255200)
rgb(0.71372549)=(0.28892100,0.75839400,0.42842600)
rgb(0.71764706)=(0.29647900,0.76156100,0.42422300)
rgb(0.72156863)=(0.30414800,0.76470400,0.41994300)
rgb(0.72549020)=(0.31192500,0.76782200,0.41558600)
rgb(0.72941176)=(0.31980900,0.77091400,0.41115200)
rgb(0.73333333)=(0.32779600,0.77398000,0.40664000)
rgb(0.73725490)=(0.33588500,0.77701800,0.40204900)
rgb(0.74117647)=(0.34407400,0.78002900,0.39738100)
rgb(0.74509804)=(0.35236000,0.78301100,0.39263600)
rgb(0.74901961)=(0.36074100,0.78596400,0.38781400)
rgb(0.75294118)=(0.36921400,0.78888800,0.38291400)
rgb(0.75686275)=(0.37777900,0.79178100,0.37793900)
rgb(0.76078431)=(0.38643300,0.79464400,0.37288600)
rgb(0.76470588)=(0.39517400,0.79747500,0.36775700)
rgb(0.76862745)=(0.40400100,0.80027500,0.36255200)
rgb(0.77254902)=(0.41291300,0.80304100,0.35726900)
rgb(0.77647059)=(0.42190800,0.80577400,0.35191000)
rgb(0.78039216)=(0.43098300,0.80847300,0.34647600)
rgb(0.78431373)=(0.44013700,0.81113800,0.34096700)
rgb(0.78823529)=(0.44936800,0.81376800,0.33538400)
rgb(0.79215686)=(0.45867400,0.81636300,0.32972700)
rgb(0.79607843)=(0.46805300,0.81892100,0.32399800)
rgb(0.80000000)=(0.47750400,0.82144400,0.31819500)
rgb(0.80392157)=(0.48702600,0.82392900,0.31232100)
rgb(0.80784314)=(0.49661500,0.82637600,0.30637700)
rgb(0.81176471)=(0.50627100,0.82878600,0.30036200)
rgb(0.81568627)=(0.51599200,0.83115800,0.29427900)
rgb(0.81960784)=(0.52577600,0.83349100,0.28812700)
rgb(0.82352941)=(0.53562100,0.83578500,0.28190800)
rgb(0.82745098)=(0.54552400,0.83803900,0.27562600)
rgb(0.83137255)=(0.55548400,0.84025400,0.26928100)
rgb(0.83529412)=(0.56549800,0.84243000,0.26287700)
rgb(0.83921569)=(0.57556300,0.84456600,0.25641500)
rgb(0.84313725)=(0.58567800,0.84666100,0.24989700)
rgb(0.84705882)=(0.59583900,0.84871700,0.24332900)
rgb(0.85098039)=(0.60604500,0.85073300,0.23671200)
rgb(0.85490196)=(0.61629300,0.85270900,0.23005200)
rgb(0.85882353)=(0.62657900,0.85464500,0.22335300)
rgb(0.86274510)=(0.63690200,0.85654200,0.21662000)
rgb(0.86666667)=(0.64725700,0.85840000,0.20986100)
rgb(0.87058824)=(0.65764200,0.86021900,0.20308200)
rgb(0.87450980)=(0.66805400,0.86199900,0.19629300)
rgb(0.87843137)=(0.67848900,0.86374200,0.18950300)
rgb(0.88235294)=(0.68894400,0.86544800,0.18272500)
rgb(0.88627451)=(0.69941500,0.86711700,0.17597100)
rgb(0.89019608)=(0.70989800,0.86875100,0.16925700)
rgb(0.89411765)=(0.72039100,0.87035000,0.16260300)
rgb(0.89803922)=(0.73088900,0.87191600,0.15602900)
rgb(0.90196078)=(0.74138800,0.87344900,0.14956100)
rgb(0.90588235)=(0.75188400,0.87495100,0.14322800)
rgb(0.90980392)=(0.76237300,0.87642400,0.13706400)
rgb(0.91372549)=(0.77285200,0.87786800,0.13110900)
rgb(0.91764706)=(0.78331500,0.87928500,0.12540500)
rgb(0.92156863)=(0.79376000,0.88067800,0.12000500)
rgb(0.92549020)=(0.80418200,0.88204600,0.11496500)
rgb(0.92941176)=(0.81457600,0.88339300,0.11034700)
rgb(0.93333333)=(0.82494000,0.88472000,0.10621700)
rgb(0.93725490)=(0.83527000,0.88602900,0.10264600)
rgb(0.94117647)=(0.84556100,0.88732200,0.09970200)
rgb(0.94509804)=(0.85581000,0.88860100,0.09745200)
rgb(0.94901961)=(0.86601300,0.88986800,0.09595300)
rgb(0.95294118)=(0.87616800,0.89112500,0.09525000)
rgb(0.95686275)=(0.88627100,0.89237400,0.09537400)
rgb(0.96078431)=(0.89632000,0.89361600,0.09633500)
rgb(0.96470588)=(0.90631100,0.89485500,0.09812500)
rgb(0.96862745)=(0.91624200,0.89609100,0.10071700)
rgb(0.97254902)=(0.92610600,0.89733000,0.10407100)
rgb(0.97647059)=(0.93590400,0.89857000,0.10813100)
rgb(0.98039216)=(0.94563600,0.89981500,0.11283800)
rgb(0.98431373)=(0.95530000,0.90106500,0.11812800)
rgb(0.98823529)=(0.96489400,0.90232300,0.12394100)
rgb(0.99215686)=(0.97441700,0.90359000,0.13021500)
rgb(0.99607843)=(0.98386800,0.90486700,0.13689700)
rgb(1.00000000)=(0.99324800,0.90615700,0.14393600)},
}
\usepackage{subcaption}

\title{A Deep Reinforcement Learning Approach to Rare Event Estimation}
\author{
    Anthony Corso, \textsuperscript{\rm 1}
    Kyu-Young Kim, \textsuperscript{\rm 1}
    Shubh Gupta, \textsuperscript{\rm 1}
    Grace Gao, \textsuperscript{\rm 1}
    Mykel J. Kochenderfer, \textsuperscript{\rm 1}
}
\affiliations{
    \textsuperscript{\rm 1} Stanford University\\
    \{acorso, kykim153, shubhgup, gracegao, mykel\}@stanford.edu
    
}

\begin{document}

\maketitle

\begin{abstract}
An important step in the design of autonomous systems is to evaluate the probability that a failure will occur. In safety-critical domains, the failure probability is extremely small so that the evaluation of a policy through Monte Carlo sampling is inefficient. Adaptive importance sampling approaches have been developed for rare event estimation but do not scale well to sequential systems with long horizons. In this work, we develop two adaptive importance sampling algorithms that can efficiently estimate the probability of rare events for sequential decision making systems. The basis for these algorithms is the minimization of the Kullback-Leibler divergence between a state-dependent proposal distribution and a target distribution over trajectories, but the resulting algorithms resemble policy gradient and value-based reinforcement learning. We apply multiple importance sampling to reduce the variance of our estimate and to address the issue of multi-modality in the optimal proposal distribution. We demonstrate our approach on a control task with both continuous and discrete actions spaces and show accuracy improvements over several baselines.

\end{abstract}

\section{Introduction}
Autonomous systems for safety-critical applications such as ground transportation, aviation, manufacturing, and in-home robotics have the potential to improve both safety and efficiency. Before an autonomous system is deployed, however, its safety needs to be rigorously evaluated in simulation. A common evaluation metric is the probability that the system fails, a task known as \emph{rare event estimation}. While definitions of failure are domain specific, the challenges of computing the probability of failure are shared across applications. For well-designed systems, failure events will be extremely rare, typically making Monte Carlo (MC) sampling-based approaches inefficient. Additionally, modern autonomous systems are designed with complex components (e.g deep neural networks) and operate in complex and nonlinear environments, which makes analytical approaches intractable. For these reasons, many safety evaluation approaches treat the autonomous system as a black box and rely on adaptive sampling of the environment to find failures~\cite{corso2021survey}. The high dimensional state spaces and long time horizons of many applications lead to a vast search space of possible trajectories. 

\begin{figure}[t]
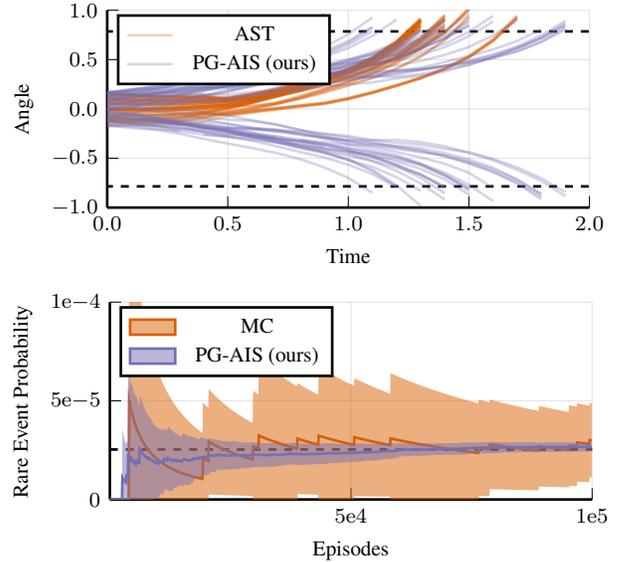

     \centering
     \begin{subfigure}[b]{\columnwidth}
         \centering
         \input{figures/trajectories}
     \end{subfigure}%
     \vspace{-0.5em}
     \begin{subfigure}[b]{\columnwidth}
         \centering
         \input{figures/estimation}
     \end{subfigure}
    \caption{\textbf{Top:} Failure trajectories from an inverted pendulum system. AST~\cite{lee2020adaptive} focuses on a single failure mode while our approach approximates the true distribution over failures. \textbf{Bottom:} Estimates of the probability of failure (shaded region is standard error) showing that our approach is more accurate with fewer samples than the MC baseline.}
    \label{fig:intro}
\end{figure}

Prior work on safety evaluation of sequential systems has focused on falsification, where the goal is to find individual failure trajectories. Approaches based on optimization~\cite{deshmukh2017testing,aerts2018temporal}, trajectory-planning~\cite{tuncali2019rapidly,zutshi2014multiple} and reinforcement learning~\cite{Akazaki2018falsification,kuutti2020training} have been applied in the context of autonomous driving~\cite{abeysirigoonawardena2019generating,corso2019adaptive} and aviation~\cite{delmas2019evaluation,julian2020validation}. Adaptive stress testing (AST)~\cite{lee2020adaptive} incorporates the likelihood of the trajectory in the cost function so as to find the most-likely failure. The downside to falsification approaches, however, is that they may converge on a single failure mode (e.g. the most likely, the most egregious, or the first one discovered) and do not adequately explore the space of failures (see 
AST example in the top plot of \cref{fig:intro}). \citet{rose2021reinforcement} also apply reinforcement learning algorithms to sample rare event trajectories, but rely on the ability to arbitrarily choose state transitions.

Algorithms for rare event estimation have used importance sampling~\cite{owen2000safe} and its variants~\cite{bugallo2017adaptive} to efficiently estimate the probability that a rare event occurs. These approaches rely on the ability to sample all stochastic variables from a proposal distribution, which poses two problems for sequential problems. First, many sequential problems have stochastic transitions that aren't directly controllable. Second, the dimensionality of the proposal scales with the time horizon, making even simple problems intractable when the horizon is long. Importance sampling approaches have been used to evaluate sequential systems by controlling a small set of fixed parameters~\cite{uesato2019rigorous,kim2016improving}, but have only been extended to sequential sampling in the form of dynamic programming algorithms~\cite{Chryssanthacopoulos2010,corso2020scalable} that do not scale to larger state spaces. 

In this work, we develop two rare event estimation algorithms for black-box sequential systems. The basis for our approach is to adaptively learn a state-dependent proposal distribution (proposal policy) by minimizing the Kullback-Leibler (KL) divergence between the distribution of trajectories induced by the proposal policy and a target distribution. The two algorithms we derive resemble policy gradient and value-based reinforcement learning, so we borrow techniques from the deep reinforcement learning literature to improve training stability and sample efficiency. Unlike reward maximization, however, efficient rare event estimation often requires a multi-modal proposal. To achieve multi-modality, and reduce the variance of our final estimate, we apply multiple importance sampling by learning a family of proposal policies and their mixture weights. We demonstrate our approaches on the evaluation of a inverted pendulum controller and show efficiency improvements over a baseline (see bottom of \cref{fig:intro}). Our contributions are as follows.
\begin{itemize}
    \item We develop two reinforcement learning-based rare event estimation algorithms for sequential problems
    \item We demonstrate those algorithms on a simulated environment with both continuous and discrete actions, showing efficiency improvements over baselines.
    \item We investigate the effect of multiple importance sampling techniques for capturing multi-modal behavior and variance reduction.
\end{itemize}

\section{Preliminaries}
We review techniques for efficient rare event estimation using importance sampling and its variants. Then we discuss rare event estimation in the context of sequential decision making systems and highlight some of the challenges to be addressed by our proposed approach.

\subsection{Rare Event Estimation}

Suppose we have a random variable $X$ with probability density $p(x)$, a function $f(x)$ and a threshold $\gamma$ such that $\mathbb{P}\left[ f(X) > \gamma \right]$ is small.  Rare event estimation is the problem of computing an accurate estimate of this probability in the form of an expectation
\begin{equation}
\mu = \mathbb{E}\left[ \I{f(X) >  \gamma} \right]
\end{equation}
where $\I{\cdot}$ is the indicator function. Monte Carlo estimation uses $p$ to generate $N$ independent samples to compute an empirical estimate
\begin{equation}
    \hat{\mu}_p = \frac{1}{N} \sum_{i=1}^N \I{f(x_i) > \gamma}
\end{equation} 
The relative accuracy of this estimator can be expressed by the coefficient of variation $c_v$, which can be shown to be
\begin{equation}
c_v(\hat{\mu}) = \sqrt{\frac{(1-\mu)}{\mu N}}
\end{equation}
The number of samples required for a target relative accuracy $c_v < \epsilon_{\rm rel}$ is therefore
\begin{equation}
    N > \frac{(1 - \mu)}{\mu  \epsilon^2_{\rm rel} }
\end{equation}
In safety-critical scenarios, $\mu$ may be as low as \num{e-9} and would require \num{e11} samples to get a relative error of less than \num{10}\%.

\paragraph{Importance Sampling} To reduce the variance of the estimator, samples may be drawn from a proposal distribution with probability density $q(x)$. The importance sampling estimate is 
\begin{equation}
    \hat{\mu}_q = \frac{1}{N} \sum_{i=1}^N w(x_i) \I{f(x_i) > \gamma}\label{eq:IS_estimation}
\end{equation}
where the samples are reweighted according to the importance weight
\begin{equation}
    w(x) = p(x)/q(x)
\end{equation}
The estimator $\hat{\mu}_q$ is unbiased if $q(x) > 0$ everywhere $\I{f(x_i) > \gamma}p(x) > 0$ and its variance is minimized by 
\begin{equation}
    q^*(x) = \frac{\I{f(x) > \gamma} p(x)}{\mu}
\end{equation}
meaning that samples should be drawn only in the region where $f(x) > \gamma$ and with likelihood proportional to the probability they occur under the nominal distribution. The challenge in importance sampling is choosing the proposal distribution $q$, since a poor choice can lead to estimators with high variance.

\paragraph{Adaptive Importance Sampling}
Adaptive importance sampling (AIS) algorithms~\cite{bugallo2017adaptive} iteratively adapt a proposal $q$ toward the optimal proposal $q^*$, eliminating the need to choose a good proposal a priori. Many AIS approaches~\cite{cappe2004population,cappe2008adaptive,martino2017layered,elvira2017improving,sinha2020neural} work by iteratively adapting a population of samples $\{x_1, \ldots, x_N \}$, while others~\cite{rubinstein2004cross,cornuet2012adaptive,Martino2015adaptive} focus on iteratively estimating the parameters $\theta$ of a proposal distribution $q_\theta$. In sequential decision-making problems, the domain of our samples (full trajectories) is too complex to adapt the samples directly. Instead, we focus on estimating the parameters of a proposal distribution. In particular we  build upon the cross-entropy method (CEM)~\cite{rubinstein2004cross} for adaptive rare event estimation. 

At each iteration $k$ the CEM updates a parameterized proposal distribution $q_{\theta_k}$ to $q_{\theta_{k+1}}$ by minimizing the Kullback-Leibler (KL) divergence $D_{\rm KL}\infdivx{\cdot}{\cdot}$ to a sample estimate of the optimal proposal distribution
\begin{align}
    \theta_{k+1} &= \argmin_{\theta} D_{\rm KL}\infdivx{q^*}{q_\theta} \\
    &\approx \argmax_\theta \frac{1}{N} \sum_{i=1}^N \I{f(x_i) > \gamma} w(x_i) \log q_\theta(x_i) \label{eq:CEM_update}
\end{align}
When failure events are rare, there may be too few failure samples to provide a good approximation to the target distribution. One solution is to adaptively choose the target threshold $\gamma_k$ at each iteration $k$ to ensure that a minimum fraction $\rho$ of samples are considered failures. The resulting algorithm takes the following steps at each iteration
\begin{enumerate}
    \item Sample $\{x_i, \ldots, x_N\}$ from $q_{\theta_k}$ with weights $w(x_i)$
    \item Sort the samples by $f(x_i)$
    \item Set $\gamma_k = \max(\gamma, f(x_{\rho N}))$
    \item Set $\theta_{k+1}$ according to \cref{eq:CEM_update}
\end{enumerate}
All samples and their importance weights are then used to estimate $\mu$ according to \cref{eq:IS_estimation}.

\paragraph{Multiple Importance Sampling}
If the family of proposal distributions described by $q_\theta$ is not well suited to approximate the optimal proposal distribution, even AIS approaches may lead to high variance estimates of $\mu$. To mitigate this problem, multiple importance sampling (MIS) uses samples from a set of $M$ proposal distributions $\{q^{(1)}, \ldots, q^{(M)} \}$. While the standard importance weighting scheme can be used (where each sample is weighted according to the proposal it was sampled from), the technique of deterministic mixture (DM) weighting~\cite{owen2000safe}, where
\begin{equation}
    w_{\rm DM}(x_i) = \frac{p(x_i)}{\frac{1}{M} \sum_{m=1}^M q^{(m)}(x_i)}\label{eq:dm_weights}
\end{equation}
has provably lower variance~\cite{elvira2019generalized} at the cost of higher computation. When one of the proposal distributions is the nominal distribution $p$, then the technique is known as defensive importance sampling~\cite{hesterberg1995weighted}, which comes with an upper bound on the worst case variance of the estimator.

\subsection{Rare Event Estimation for MDP Polices}
When developing safety-critical autonomous systems it is crucial to compute the probability that rare but catastrophic failures will occur. This section reviews sequential decision making systems and how importance sampling can be applied to their evaluation.

\paragraph{MDPs and Policies}
A Markov Decision Process (MDP) is a model for sequential decision making defined by ($S$, $A$, $R$, $P$) where $S$ is the state space, $A$ is the action space, $R$ is the reward function and $P$ is the transition function. An agent observes the current state $s$ and chooses an action $a$ causing the system to transition to a new state $s^\prime$ with probability $P(s^\prime \mid s, a)$, then receives a reward $r=R(s, a)$. 

A policy $\pi(a \mid s)$ is a function that maps a state to a distribution over actions. In this work, we assume that the MDP is finite-horizon, meaning rollouts of a policy produces state-action trajectories trajectories $\tau = \{s_0, a_1, s_1, \ldots, a_T, s_T\}$ with finite length. Policy evaluation is the process of computing the expected sum of future rewards
\begin{equation}
    V^{\pi}(s) = \mathbb{E}\left[ R(\tau) \mid s_0 = s\right]\label{eq:value_function}
\end{equation}
where $R(\tau) = \sum_{t=1}^T R(s_t, a_t)$ is the return of a trajectory and the expectation is taken over the distribution induced by the policy $\pi(a_t \mid s_t)$ and transition function $P(s_{t+1} \mid s_t, a_t)$. If the state and action spaces are small enough, policies can be evaluated through exact or approximate dynamic programming, but for larger environments or environments that are strictly episodic (i.e., they cannot be initialized into arbitrary states), sampling and online learning is required~\cite{kochenderfer2022algorithms}. In this work, we design our approach for episodic simulators. 

\paragraph{Importance Sampling in MDPs}
Rare event estimation for sequential problems involves accurately estimating
\begin{equation}
    \mu = \mathbb{E}\left[ \I{R(\tau) >\gamma} \right]
\end{equation}
Following the theory of importance sampling, the optimal proposal distribution is
\begin{equation}
    q^*(\tau) = \frac{\I{R(\tau) >\gamma} p(\tau)}{\mu}
\end{equation}
We cannot, however, naively apply the previous importance sampling techniques because 1) the domain of feasible trajectories is too complex to be represented by a parameterized distribution and 2) we have no way to evaluate the probability density of a trajectory 
\begin{equation}
    p(\tau) = P(s_0) \prod_{t=1}^T P(s_{t} \mid s_{t-1}, a_t) \pi(a_t \mid s_{t-1})
\end{equation}
because it depends on the transition function $P$ which can only be sampled from. Instead, we must learn a proposal distribution over the set of variables we have control over: the actions. 

Let $q_\theta(a \mid s)$ be a \emph{proposal policy} parameterized by $\theta$. Trajectories sampled from rollouts of this policy have density
\begin{equation}
q_\theta(\tau) = P(s_0) \prod_{t=1}^T P(s_{t} \mid s_{t-1}, a_t) q_\theta(a_t \mid s_{t-1})\label{eq:proposal_density}
\end{equation}
The importance weight of these samples is independent of the transition function since those terms cancel out and give
\begin{equation}
    w(\tau) = \frac{p(\tau)}{q_\theta(\tau)} = \frac{ \prod_{t=1}^T \pi(a_t \mid s_{t-1})}{\prod_{t=1}^T q_\theta(a_t \mid s_{t-1})}\label{eq:sequential_importance_weight}
\end{equation}
For the derivation of the algorithms in the next section, we also define a partial importance weight for a state $s_k$ as
\begin{equation}
    w(s_k) = \frac{ \prod_{t=1}^{k-1} \pi(a_t \mid s_{t-1})}{\prod_{t=1}^{k-1} q_\theta(a_t \mid s_{t-1})}\label{eq:partial_sequential_importance_weight}
\end{equation}
where all states and actions come from the same trajectory.


\paragraph{Adversarial MDP Formulation}
Prior work~\cite{corso2021thesis} has shown that the variance of an importance sampling estimator for a sequential problem has an inverse relationship with the amount of influence the actions have over the state transitions. Also, many policies we may wish to evaluate are deterministic, so all of the stochasticity comes from the transition function. For these reasons, safety evaluation can be formulated as an adversarial MDP~\cite{corso2021survey}. 

An adversarial MDP is constructed from an original MDP and a policy. It has the same state space as the original MDP but the actions are replaced with variables that control the stochasticity of the original transition function and have a known distribution $\pi(a \mid s)$. The actions of a deterministic policy, then, are subsumed into the transition function of the adversarial MDP. The reward function is replaced with a risk metric where high returns imply rare events of interest.

For example, prior work on the safety validation of an autonomous driving policy~\cite{koren2018adaptive} constructed an adversarial MDP where the actions controlled the motion of a pedestrian as well as the sensor noise experienced by the autonomous vehicle, the transition function combined the driving simulator dynamics with the vehicle's actions, and a reward was given when the vehicle collided with the pedestrian. Stochastic models of these actions were developed using expert knowledge. 


\section{Methods}
Building upon the cross-entropy method, we apply an adaptive importance sampling algorithm that iteratively minimizes the KL-divergence between the distribution induced by the current proposal policy and the optimal target distribution. We update the parameters $\theta$ of a distribution $q_\theta$ with gradient descent as
\begin{equation}
    \theta \gets \theta - \alpha \nabla_\theta D_{\rm KL}\infdivx{q^*(\tau)}{q_\theta(\tau)}\label{eq:general_update}
\end{equation}
where $\alpha$ is the learning rate. To ensure a sufficient number of samples are used to compute the gradient, the failure threshold $\gamma$ is gradually increased to always include a fraction $\rho$ of the training samples. 

In the following subsections we explore policy gradient and value-based methods for estimating the gradient in \cref{eq:general_update}. Then we discuss how multiple importance sampling can be used to address high variance and multi-modality. 

\subsection{Policy Gradient Approach}

Our first approach is to approximate the gradient using full stochastic rollouts of the environment. If we obtain trajectories according to a proposal distribution $\tau \sim q_\varphi(\cdot)$, the gradient is estimated by
\begin{equation}
\begin{split}
    \nabla_\theta D_{\rm KL}&\infdivx{q^*(\tau)}{q_\theta(\tau)} \approx \\ &-\nabla_\theta \frac{1}{N} \sum_{i=1}^N  \I{R(\tau_i) >\gamma} w(\tau_i) \log q_\theta(\tau_i) 
\end{split}
\end{equation}
where $w(\tau)$ is given by \cref{eq:sequential_importance_weight}. This expression can be simplified by using the fact that $\theta$ only controls the distribution over actions, and not the distribution of environment transitions. Take the $\log$ of \cref{eq:proposal_density} to separate out terms that depend on $\theta$ and terms that do not
\begin{equation}
\begin{split}
    \log q_\theta(\tau) = \log p(s_0) + &\sum_{t=1}^T \log p(s_{t+1} \mid s_{t}, a_{t}) + \\ 
    &\sum_{t=1}^T \log q_\theta(a_{t} \mid s_{t})
\end{split}
\end{equation}
Plugging this into the expression, and dropping all terms that don't depend on $\theta$, we have
\begin{equation}
\begin{split}
    &\nabla_\theta D_{\rm KL}\infdivx{q^*(\tau)}{q_\theta(\tau)} \approx\\
    &-\nabla_\theta \frac{1}{N} \sum_{i=1}^N \sum_{t=1}^T  \I{R(\tau_i) >\gamma} w(\tau_i)  \log q_\theta(a_{t,i} \mid s_{t,i})  \label{eq:rollout_gradient}
\end{split}
\end{equation}
The expression in \cref{eq:rollout_gradient} is now in a form that can be minimized using stochastic gradient decent.

\paragraph{Variance Reduction with a Baseline}
The expression in \cref{eq:rollout_gradient} takes a similar form to the policy gradient in reinforcement learning~\cite{sutton2018reinforcement}. There are several approaches to reducing the variance of the gradient estimator that have been developed in RL. The first, called the reward-to-go trick, only considers values of the reward after the current state. However, due to the presence of  the importance weight (which depends upon the full trajectory) we cannot apply the same trick here. We can, however, apply another variance-reducing trick: subtracting a state-dependent baseline giving
\begin{equation}
\begin{split}
    &\nabla_\theta D_{\rm KL}\infdivx{q^*(\tau)}{q_\theta(\tau)} \approx \\
    &-\nabla_\theta \frac{1}{N} \sum_{i,t=1}^{N, T} \left(\I{R(\tau_i) >\gamma} w(\tau_i) - b(s_{t,i}) \right) \log q_\theta(a_{t,i} \mid s_{t,i})
\end{split}
\end{equation}

Taking inspiration from the advantage function in reinforcement learning, we can let 
\begin{align}
    b(s_t) &= \mathbb{E}_p[ \I{R(\tau) >\gamma} \mid s_t] \\
    &=\mathbb{E}_{q_\theta}[ \I{R(\tau) >\gamma} w(\tau) \mid s_t]
\end{align}

The baseline can be estimated using a parameterized model $b_\phi$ that minimizes the mean squared error to a empirical target, with the loss function
\begin{equation}
   L(\phi) = \frac{1}{N} \sum_{i=1}^N \sum_{t=1}^T \left( b_\phi(s_{t, i}) -   \I{R(\tau_i) >\gamma} w(\tau_i) \right)^2\label{eq:baseline_loss}
\end{equation}

The final algorithm, which we will call policy gradient adaptive importance sampling (PG-AIS), is shown in \cref{alg:pgais}. At each iteration, $\Delta N$ trajectories are sampled according to the current proposal policy (line~\ref{line:pgais_sample}). They are sorted into ascending order and used with the elite fraction parameters $\rho$ to compute the current threshold (line~\ref{line:pgais_threshold}. The returns and weights are then computed (lines~\ref{line:pgais_returns}--\ref{line:pgais_weights}) and stored. The most recently sampled data is put into a training buffer (line~\ref{line:pgais_training_buffer}) and used with stochastic gradient descent (SGD) to update the proposal policy (line~\ref{line:pgais_proposal}) and the baseline (line~\ref{line:pgais_baseline}). After $N_{\rm tot}$ samples are collected, the algorithm computes and returns the importance sampling estimate $\hat{\mu}$. 

\begin{algorithm}
\caption{Policy Gradient Adaptive Importance Sampling.} \label{alg:pgais}
\begin{algorithmic}[1]
    \Function{PG-AIS}{$\theta_0$, $\phi_0$, $\gamma$, $N$, $N_{\rm tot}$, $\rho$}
    \State $D \gets \emptyset$
    \State $k \gets 0$
    \While{$\textproc{Length}(D) < N_{\rm tot}$}
        \State Sample $\{ \tau_1, \ldots, \tau_{N}\}$ using $q_{\theta_k}$, then sort \label{line:pgais_sample}
        \State Compute $\gamma_k \gets \min(\gamma, \tau{\rho_{N}})$ \label{line:pgais_threshold}
        \State Compute returns $R_i \gets R(\tau_i)$\label{line:pgais_returns}
        \State Compute weights $w_i \gets w(\tau_i)$ from \cref{eq:sequential_importance_weight}\label{line:pgais_weights}
        \State $D \gets D \bigcup \{(R_i, w_i)\}_{i=1}^N$
        \State $B \gets \{(s_{t,i}, a_{t,i}, w_i, R_i)\}_{i=1}^N$  \Comment{Training buffer}\label{line:pgais_training_buffer}
        \State $\theta_{k+1} \gets $ \textproc{SGD}($\theta_k$, $B$) with \cref{eq:rollout_gradient}\label{line:pgais_proposal}
        \State $\phi_{k+1} \gets $ \textproc{SGD}($\phi_k$, $B$) with \cref{eq:baseline_loss}\label{line:pgais_baseline}
        \State $k \gets k + 1$
    \EndWhile
    \State \Return{$\hat{\mu} \gets$ \textproc{Estimate}($D$, $\gamma$)} \Comment{using \cref{eq:IS_estimation}}
    \EndFunction
\end{algorithmic}
\end{algorithm}

\subsection{Value-Based Approach}
Value-based methods, where the policy is optimized with respect to a learned value function, have been successful for improving sample efficiency in deep RL~\cite{mnih2015human}. To employ a similar approach, we first need to identify the function against which to optimize the policy. We do this by deriving the optimal proposal policy $q^*(a \mid s)$.

To simplify derivation, we make two assumptions. The first assumption that the reward function $R$ is sparse, meaning it is zero everywhere except in terminal states $S_{\rm term}$ where it can have a nonzero value. Note that any non-sparse function can be converted to a sparse function by augmenting the state space with a running total of the return. The second assumption is that the state transitions are deterministic given the action. This is a common assumption when evaluating the safety of autonomous systems using an adversarial MDP formulation~\cite{corso2021survey} but may not be true in general.  Under these assumptions, prior work~\cite{corso2020scalable} has show that the optimal proposal policy $q^*(a \mid s)$ is given by
\begin{equation}
    q^*(a \mid s) = \frac{ Q^\pi(s,a) \pi(a \mid s)}{V^\pi(s)} \label{eq:optimal_policy}
\end{equation}
where $Q^\pi(s,a) = \mathbb{E}\left[ R(\tau) \mid s_0=s, \ a_0=a \right]$ is the action value function and $V^\pi$ is given by \cref{eq:value_function}.

When the action space is discrete, the optimal policy can be computed directly from an estimate of the action-value function by enumerating the actions and computing \cref{eq:optimal_policy}. With continuous actions, however, we need to optimize the proposal policy to match the target policy. To do so, we rewrite the trajectory-level KL divergence in terms of the divergence between policies as
\begin{equation}
\begin{split}
\nabla_\theta D_{\rm KL}&\infdivx{q^*(\tau)}{q_\theta(\tau)} = \\
    &\nabla_\theta \int_{S} \nu(s) D_{\rm KL}\infdivx{q^*(a \mid s)}{q_\theta(a \mid s)} ds
\end{split}
\end{equation}
where $\nu(s)$ is the visitation frequency. Substituting \cref{eq:optimal_policy} gives
\begin{equation}
 \begin{split}
  &\nabla_\theta D_{\rm KL}\infdivx{q^*(\tau)}{q_\theta(\tau)} =\\
    &-\nabla_\theta \int_{S} \nu(s) \mathbb{E}_{q_{\theta}} \left[ \frac{ Q^\pi(s,a) \pi(a \mid s)}{q_\theta(a \mid s)V^\pi(s)} \log q_\theta(a \mid s) \right]ds
\end{split}
\end{equation}
which can be estimated with samples of the state $s$ as
\begin{equation}
\begin{split}
    &\nabla_\theta D_{\rm KL}\infdivx{q^*(\tau)}{q_\theta(\tau)} \approx \\
    &\frac{1}{N} \sum_{i=1}^N w(s_i) \mathbb{E}_{a \sim q_\theta}\left[ \frac{ Q^\pi(s_i,a) \pi(a \mid s_i)}{q_\theta(a \mid s_i)V^\pi(s_i)} \log q_\theta(a \mid s_i) \right]\label{eq:vb_gradient}
\end{split}
\end{equation}
The gradient of the expectation is computed by sampling actions with the reparameterization trick~\cite{kingma2013auto}.

With both discrete and continuous policies covered, our remaining goal is to estimate $Q^\pi$ using a parameterized function $Q^\pi_\varphi$. Under our assumptions on $R$, we can write $Q^\pi$ as a recursive formula
\begin{equation}
Q^\pi(s, a) = \left \{ \begin{array}{ll}
\I{R(s,a) >\gamma} & \quad \text{if} \ s \in S_{\rm term} \\
\mathbb{E} \left[Q^\pi(s', a') \right] & \quad \text{otherwise}
\end{array}\right.
\end{equation}
which is amenable to bootstrap updates (similar to the Bellman update in $Q$-learning~\cite{sutton2018reinforcement}). We therefore define the loss function
\begin{equation}
    L(\varphi) = \mathbb{E}\left[ (Q^\pi_\varphi(s, a) - y )^2\right]
\end{equation}
where the target $y$ is computed as
\begin{equation}
y(s) = \left \{ \begin{array}{ll}
\I{R(s,a) >\gamma} & \quad \text{if} \ s' \in S_{\rm term} \\
\mathbb{E} \left[Q^\pi_{\varphi^-}(s', a') \right] & \quad \text{otherwise}
\end{array}\right.\label{eq:target}
\end{equation}
where $\varphi^-$ is a set of stationary target parameters~\cite{mnih2015human}, and the expectation is computed exactly for discrete actions and with numerical integration techniques for continuous actions.

Because state and action samples are obtained using a proposal policy in the environment, we must estimate the loss using importance weights 
\begin{equation}
    L(\varphi) = \frac{1}{N} \sum_{i=1}^N w(s_{i+1}) (Q^\pi(s_i, a_i) - y(s_i^\prime))^2\label{eq:QLoss}
\end{equation}

The final algorithm, called value-based adaptive importance sampling (VB-AIS), is shown in \cref{alg:vbais}. The main differences from \cref{alg:pgais} is that batches are sampled from a replay buffer (line \ref{line:vbais_batch_sampled}) and used to compute a target (line \ref{line:vbais_target}) to update the value function estimate (line \ref{line:vbais_value_function}). Then the proposal policy is updated with respect to the action value function (line \ref{line:vbais_policy_update}). 

\begin{algorithm}
\caption{Value Based Adaptive Importance Sampling.} \label{alg:vbais}
\begin{algorithmic}[1]
    \Function{VB-AIS}{$\theta_0$, $\phi_0$, $\gamma$, $N$, $N_{\rm tot}$, $\rho$}
    \State $D \gets \emptyset$
    \State $k \gets 0$
    \While{$\textproc{Length}(D) < N_{\rm tot}$}
        \State Sample $\{ \tau_1, \ldots, \tau_{N}\}$ using $q_{\theta_k}$, then sort
        \State Compute $\gamma_k \gets \min(\gamma, \tau{\rho_{N}})$
        \State Compute returns $R_i \gets R(\tau_i)$
        \State Compute weights $w_i \gets w(\tau_i)$ from \cref{eq:sequential_importance_weight}
        \State $D \gets D \bigcup \{(R_i, w_i)\}_{i=1}^N$
        \State Sample batch $B$ from $D$\label{line:vbais_batch_sampled}
        \State Compute $y$ from \cref{eq:target}\label{line:vbais_target}
        \State $\varphi_{k+1} \gets $ \textproc{SGD}($\varphi_k$, $B$) with \cref{eq:QLoss}\label{line:vbais_value_function}
        \State $\theta_{k+1} \gets $ \textproc{SGD}($\theta_k$, $B$) with \cref{eq:vb_gradient}\label{line:vbais_policy_update}
        \State $\varphi_{k+1}^- \gets \varphi_{k+1}$ at desired interval
        \State $k \gets k+1$
    \EndWhile
    \State \Return{$\hat{\mu} \gets$ \textproc{Estimate}($D$, $\gamma$)} \Comment{using \cref{eq:IS_estimation}}
    \EndFunction
\end{algorithmic}
\end{algorithm}

\subsection{Multiple Importance Sampling}
There are two reasons to apply multiple importance sampling to our approaches. First, it may reduce the variance of the final importance sampling estimate. Second, if the optimal proposal distribution is multi-modal, it will not be well approximated by a single mode proposal. For continuous actions, we use a Gaussian policy, and therefore require multiple proposals to represent more complex distributions. 

Applying MIS to these algorithms is  straightforward. We use a set of $M$ proposals with parameters $\{\theta_1, \ldots, \theta_M \}$. During sampling, a predetermined number of samples are taken from each distribution and the importance weights are computed according to \cref{eq:dm_weights} instead of \cref{eq:sequential_importance_weight}. The proposals are then updated using a variant of expectation maximization. First, the samples are reassigned to the proposal that gives them the maximum likelihood (maximization step). Then each proposal (and corresponding baseline or state-action value function) is updated only with samples that have been assigned to it (expectation step). Note that we use a hard sample assignment (instead of a soft weighting) because it had little effect on the performance while being computationally more efficient.

\section{Experiments}\label{sec:experiments}

This section describes our experimental evaluation of the proposed rare event estimation algorithms. We start by introducing our environment, then describe our metrics and baselines, and lastly present and discuss the results.

\subsection{Environment}
The environment we used for evaluation is the control of an inverted pendulum. We estimate the failure rate of a nonlinear control policy (designed to balance the pendulum) subject to external disturbances, modeled as additive torques on the pendulum. We consider the case of discrete disturbance, where the ground truth failure rate is \num{2.53e-5}, and continuous disturbances, where the failure rate is \num{1.96e-5}. The inverted pendulum  has three continuous states and a trajectory length of $\num{20}$ timesteps making it sufficiently high dimensional for naive approaches to fail. Additionally, it has two distinct failure modes (falling to the left or right), which is typical of rare event estimation problems and challenging for naive algorithms. See the appendix for additional details. 

\subsection{Evaluation Metrics and Baselines}
Rare event estimation algorithms are evaluated on their error (absolute deviation) in estimating the occurrence rate $\mu$. To avoid a dependence on the magnitude of $\mu$, we divide by $\mu$ to compute the relative absolute deviation
\begin{equation}
    \epsilon_{\rm abs} = |\hat{\mu} - \mu| / \mu
\end{equation}
Additionally, we consider the relative error 
\begin{equation}
    \epsilon_{\rm rel} = (\hat{\mu} - \mu)/\mu
\end{equation}
which, when averaged across trials, gives an estimate of the empirical bias of the estimator. 

We obtain a good estimate of the ground truth value of $\mu$ with \num{5e6} Monte Carlo samples and then use it to evaluate $\epsilon_{\rm abs}$ and $\epsilon_{\rm rel}$ for each of our algorithms after a prescribed number of samples ($N_{\rm tot}=\num{5e4}$). We repeat each experiment \num{10} times and report the mean and standard deviation of the metrics.

We compare PG-AIS and VB-AIS (shortened to PG and VB) to Monte Carlo (MC) estimation and a variant of the cross entropy method (CEM). The cross entropy method learns a state-independent proposal policy from the same samples used to train the state-dependent policy of PG.


\subsection{Results}

\begin{table*}
\small
    \centering
    \begin{tabular}{@{}llll|lll@{}} 
    \toprule
     & \multicolumn{3}{c}{\textbf{Pendulum (Discrete)}} & \multicolumn{3}{c}{\textbf{Pendulum (Continuous)}}\\ 
    \midrule
    \textbf{Method} & \textbf{$M=1$} & \textbf{$M=2$} & \textbf{$M=4$} & \textbf{$M=1$} & \textbf{$M=2$} & \textbf{$M=4$}\\
    \midrule
    MC & $0.99 \pm 0.49$ & - & - & $1.03 \pm 0.68$ & - & -\\ 
    CEM & $0.51 \pm 0.14$ & $0.62 \pm 0.37$ & $0.52 \pm 0.24$ & $0.52 \pm 0.05$ & $0.25 \pm 0.21$ & $0.34 \pm 0.32$\\ 
    PG & $0.22 \pm 0.22$ & $0.18 \pm 0.15$ & $\bm{0.11 \pm 0.05}$ & $0.51 \pm 0.02$ & $\bm{0.10 \pm 0.08}$ & $0.14 \pm 0.23$\\ 
    VB & $0.43 \pm 0.25$ & $0.35 \pm 0.24$ & $0.37 \pm 0.27$ & $0.56 \pm 0.33$ & $0.30 \pm 0.20$ & $0.34 \pm 0.23$\\ 
    \bottomrule
    \end{tabular}
    \caption{Average absolute deviation $\bar{e}_{abs}$. $M$ is the number of proposals.}
    \label{tab:acc}
\end{table*}

\begin{table*}
\small
    \centering
    \begin{tabular}{@{}llll|lll@{}} 
    \toprule
     & \multicolumn{3}{c}{\textbf{Pendulum (Discrete)}} & \multicolumn{3}{c}{\textbf{Pendulum (Continuous)}}\\ 
    \midrule
    \textbf{Method} & \textbf{$M=1$} & \textbf{$M=2$} & \textbf{$M=4$} & \textbf{$M=1$} & \textbf{$M=2$} & \textbf{$M=4$}\\
    \midrule
    MC & $-0.45 \pm 1.06$ & - & - & $0.43 \pm 1.20$ & - & -\\ 
    CEM & $-0.51 \pm 0.14$ & $-0.62 \pm 0.37$ & $-0.50 \pm 0.29$ & $-0.52 \pm 0.05$ & $-0.25 \pm 0.21$ & $-0.05 \pm 0.48$\\ 
    PG & $0.13 \pm 0.29$ & $0.18 \pm 0.15$ & $0.11 \pm 0.05$ & $-0.51 \pm 0.02$ & $-0.04 \pm 0.12$ & $0.02 \pm 0.27$\\ 
    VB & $-0.30 \pm 0.41$ & $0.06 \pm 0.44$ & $0.27 \pm 0.37$ & $-0.56 \pm 0.33$ & $-0.19 \pm 0.31$ & $-0.20 \pm 0.37$\\
    \bottomrule
    \end{tabular}
    \caption{Empirical bias $\bar{\epsilon}_{\rm rel}$. $M$ is the number of proposals.}
    \label{tab:bias}
\end{table*}

\paragraph{Overview}
The average absolute deviation and empirical bias for all algorithms across both environments are reported in \cref{tab:acc} and \cref{tab:bias}, respectively. The MC approach had the highest error, but could be improved by applying a state-independent version of the CEM. The CEM method still had large errors as well as a consistent negative bias (except for the continuous setting with \num{4} distributions). This bias is due to either 1) the missing of one of the two failure modes, or 2) the need to use a wider proposal distribution to induce failures, where many of the sampled failures have small importance weights. The PG approach had the lowest absolute deviation and bias overall (specifically with $M=4$ with discrete actions and $M=2$ with continuous actions). The VB approach showed improvement over the CEM but had higher absolute deviation and higher bias than the PG approach. 

We found that the PG and VB approaches had improved performance if we pre-trained the policies to be near $\pi(a \mid s)$ and trained the value functions to a low constant value. Additionally we found that neither the use of a learned baseline function $b$, nor the use of defensive importance sampling improved the performance of the algorithms were therefore not used to produce the final results. Ablation studies showing these three phenomena are discussed in the appendix.

\begin{figure}[t]
     \centering
     \input{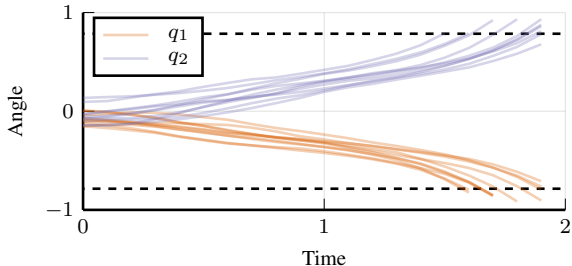}
    \caption{Demonstration of failure mode specialization for the continuous pendulum with two proposals.}
    \label{fig:mis}
\end{figure}

\paragraph{Bias Reduction with MIS}
We found the use of multiple importance sampling led to substantial improvements for the policy gradient approaches. A single proposal in the continuous action setting tended to discover only a single mode leading to large error and a consistent negative bias. Incorporating additional proposals allowed each to specialize to a different failure mode, as shown in \cref{fig:mis}. MIS also helped the CEM and the VB methods, although to a lesser extent.

\begin{figure}[t]
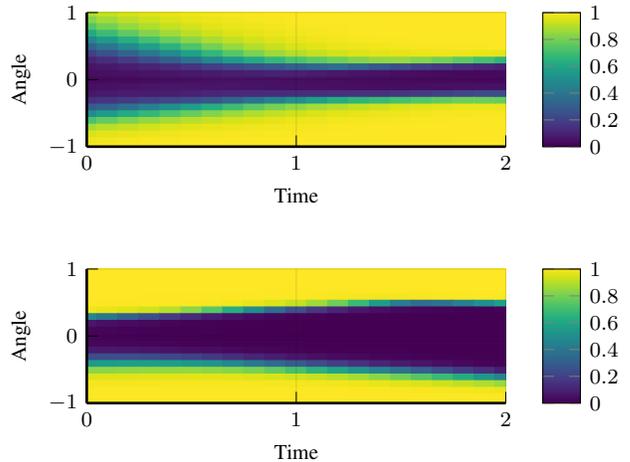

     \centering
     \begin{subfigure}[b]{\columnwidth}
         \centering
         \input{figures/mode1}
     \end{subfigure}%
     \vspace{-0.5em}
     \begin{subfigure}[b]{\columnwidth}
         \centering
         \input{figures/mode2}
     \end{subfigure}
    \caption{State-action value functions for $VB$ with \num{2} proposals on the continuous pendulum.}
    \label{fig:vb_failure}
\end{figure}

\paragraph{Challenges with VB-AIS}
One reason that the VB approach may have been less effective than the PG approach was the lack of specialization of proposals (as shown by the state-action value functions in \cref{fig:vb_failure}). Likely due to the use of a replay buffer, both proposals attempted to cover both failure modes, leading to less stable training and a more challenging policy optimization problem. Future work should investigate variations on the EM algorithm that would enable mode specialization in the value-based approach.

\paragraph{Action Space Limitations}
While these results demonstrate that the presented algorithms have potential, demonstrations on more complex environments are still required. Initial experimentation showed that both approaches fail when used with high dimensional action spaces. Approximating complex distributions in high dimensional problems is challenging, and future work should investigate solutions to this problem including the use of different proposal policy parameterizations beyond simple Gaussians.  



\section{Conclusion}
In this work we considered the problem of rare event estimation in sequential decision making systems, which is important in the evaluation of safety-critical autonomy. We extended an existing adaptive importance sampling algorithm, the cross entropy method, to the domain of sequential decisions. The resulting algorithms resembled policy gradient and value-based reinforcement learning. We demonstrated the benefit of these algorithms on a simulated control task and discussed the weaknesses of the approach that will be addressed in future work.

\bibliography{references}


\appendix

\section{Pendulum Environment}
The inverted pendulum environment has three continuous states $s = [t, \theta, \omega]$ where $t$ is the time $\theta$ is the angle of the pendulum from the vertical, and $\omega$ is the angular velocity. The discrete-time transition function for the inverted pendulum are
\begin{equation}
    \begin{split}
        \theta_{t+1} & = \theta_t + \omega_t \Delta t \\
        \omega_{t+1} & = \omega_t - \frac{3g}{2\ell}\sin(\theta_t + \pi) + \frac{3a}{m\ell^2}\Delta t
    \end{split}
\end{equation}
where $g$ is the acceleration due to gravity, $\ell$ is the length of the pendulum, $m$ is the mass of the pendulum, $\Delta t$ is the time step, and $a$ is the input torque from the controller. In our setup, we use 
$g = $\SI{10}{\meter\per\second}, $\ell = $\SI{1}{\meter}, $m = $\SI{1}{\kilogram}, and $\Delta t = $\SI{0.05}{\second}. We clip $\omega$ such that the magnitude of the angular velocity does not exceed \SI{8}{\radian\per\second}, and we clip the control inputs so that the maximum torque magnitude does not exceed \SI[inter-unit-product =\ensuremath{\cdot}]{2}{\newton\meter}.

We analyze a simple rule-based policy for balancing the under-actuated pendulum. The control rules are given by
\begin{equation}
    \begin{split}
        \omega_\text{target} & = \text{sign}(\theta) \sqrt{60(1 - \cos(\theta)} \\
        a & = -2\omega + (\omega - \omega_\text{target})
    \end{split}
\end{equation}
where the first equation determines the angular velocity required to move the pendulum from its current angle to an angle of zero. The second equation performs proportional control using this quantity and the current angular velocity.

For the pendulum with discrete disturbances, we applied torques in the set $[-1.0, 0.25, 0.0, 0.25, 1.0]$ with probabilities $[0.016, 0.30, 0.37, 0.30, 0.016]$. For the continuous disturbance case, the disturbances were sampled from $\mathcal{N}(0, 0.4)$. 

\section{Other Variance Reduction Methods}

\begin{table*}
\small
    \centering
    \begin{tabular}{@{}lll|ll@{}} 
    \toprule
     & \multicolumn{2}{c}{\textbf{Pendulum (Discrete)}} & \multicolumn{2}{c}{\textbf{Pendulum (Continuous)}}\\ 
    \midrule
    \textbf{Method} & \textbf{No Pretrain} & \textbf{Pretrain} & \textbf{No Pretrain} & \textbf{Pretrain}\\
    \midrule
    PG & $0.41 \pm 0.14$ & $0.22 \pm 0.22$ & $0.46 \pm 0.15$ & $0.51 \pm 0.02$\\ 
    PG MIS-2 & $0.17 \pm 0.08$ & $0.18 \pm 0.15$ & $0.18 \pm 0.18$ & $0.10 \pm 0.08$\\ 
    PG MIS-4 & $0.15 \pm 0.11$ & $0.11 \pm 0.05$ & $0.23 \pm 0.28$ & $0.14 \pm 0.23$\\ 
    VB & $0.70 \pm 0.70$ & $0.43 \pm 0.25$ & $0.78 \pm 0.14$ & $0.56 \pm 0.33$\\ 
    VB MIS-2 & $0.80 \pm 0.48$ & $0.35 \pm 0.24$ & $0.60 \pm 0.26$ & $0.30 \pm 0.20$\\ 
    VB MIS-4 & $0.66 \pm 0.40$ & $0.37 \pm 0.27$ & $0.86 \pm 1.27$ & $0.34 \pm 0.23$\\ 
    \bottomrule
    \end{tabular}
    \caption{Effect of policy pre-training.}
    \label{tab:pretrain}
\end{table*}

During the course of experimentation we explored a number of modifications to the baseline algorithms and report the ablations here. First, we explored the use of policy pre-training, where the proposal policy is first trained to be close the nominal policy $\pi(a \mid s)$, leading to importance weights close to \num{1} in the early part of training. The results in \cref{tab:pretrain} show that in most cases the pre-training improves performance, so we recommend its use. One exception may be in problems where significant exploration is required beyond the gradual increase of the failure threshold parameter $\gamma$, then having a higher variance proposal distribution may be helpful.

\begin{table*}
\small
    \centering
    \begin{tabular}{@{}lll|ll@{}} 
    \toprule
     & \multicolumn{2}{c}{\textbf{Pendulum (Discrete)}} & \multicolumn{2}{c}{\textbf{Pendulum (Continuous)}}\\ 
    \midrule
    \textbf{Method} & \textbf{Vanilla} & \textbf{Defensive} & \textbf{Vanilla} & \textbf{Defensive}\\
    \midrule
    PG & $0.22 \pm 0.22$ & $0.37 \pm 0.17$ & $0.51 \pm 0.02$ & $0.50 \pm 0.06$\\ 
    PG MIS-2 & $0.18 \pm 0.15$ & $0.32 \pm 0.16$ & $0.10 \pm 0.08$ & $0.21 \pm 0.21$\\ 
    PG MIS-4 & $0.11 \pm 0.05$ & $0.11 \pm 0.08$ & $0.14 \pm 0.23$ & $0.14 \pm 0.14$\\ 
    VB & $0.43 \pm 0.25$ & $0.60 \pm 0.22$ & $0.56 \pm 0.33$ & $0.65 \pm 0.21$\\ 
    VB MIS-2 & $0.35 \pm 0.24$ & $0.45 \pm 0.32$ & $0.30 \pm 0.20$ & $0.35 \pm 0.21$\\ 
    VB MIS-4 & $0.37 \pm 0.27$ & $0.23 \pm 0.18$ & $0.34 \pm 0.23$ & $0.30 \pm 0.13$\\ 
    \bottomrule
    \end{tabular}
    \caption{Effect of defensive importance sampling.}
    \label{tab:defensive}
\end{table*}

Next we explored the use of defensive sampling where the nominal distribution is included as one of the importance sampling distributions. The results in \cref{tab:defensive} show that defensive importance sampling does improve the algorithms performance. The may be because 1) the nominal distribution never samples a failure, so it contributes nothing to the overall estimate, and 2) the proposal policies are not so different from the nominal, they they require the weight smoothing that defensive importance sampling offers.

\begin{table*}
\small
    \centering
    \begin{tabular}{@{}lll|ll@{}} 
    \toprule
     & \multicolumn{2}{c}{\textbf{Pendulum (Discrete)}} & \multicolumn{2}{c}{\textbf{Pendulum (Continuous)}}\\ 
    \midrule
    \textbf{Method} & \textbf{No Baseline} & \textbf{Baseline} & \textbf{No Baseline} & \textbf{Baseline}\\
    \midrule
    PG & $0.22 \pm 0.22$ & $0.88 \pm 0.75$ & $0.51 \pm 0.02$ & $0.99 \pm 0.02$\\ 
    PG MIS-2 & $0.18 \pm 0.15$ & $0.46 \pm 0.26$ & $0.10 \pm 0.08$ & $0.93 \pm 0.14$\\ 
    PG MIS-4 & $0.11 \pm 0.05$ & $0.54 \pm 0.27$ & $0.14 \pm 0.23$ & $0.96 \pm 0.09$\\ 
    \bottomrule
    \end{tabular}
    \caption{Effect of baseline for policy gradient algorithms.}
    \label{tab:baseline}
\end{table*}

Lastly, we explored the use of a baseline for the policy gradient variants, with the hypothesis that it may improve training stability, and therefore algorithm performance. The results in \cref{tab:baseline} show that the use of the baseline actually worsened performance of the PG algorithms. The baseline values that are being estimated are all very small numbers (less than the failure rate), which may be a challenge for the neural network representation. If the learned baseline doesn't represent the target values well, then the variance of gradient estimate might increase due to its presence, which appears to be happening here. We therefore do not recommend using a baseline, except maybe in problems that have significant stochasticity in the state transitions, that is unaccounted for by the actions. 

\section{Hyperparameter Values and Training}
All experiments were done using single-thread CPUs (intel i5 and Mac M1), though the code was parallelized across experiments.

Below are the hyperparmeters used for training.

\noindent Shared across all algorithms:
\begin{itemize}
    \item NN Architectures: \num{2} hidden layers of \num{32} units, relu activations
    \item Optimizer: Adam with learning rate \num{3e-4}
    \item Training batch size: \num{1024}
    \item Gradient clipping to $1.0$
    \item Pretraining: \num{100} epochs with \num{1e4} data points
    \item Target value for pretraining value functions: $0.1$
\end{itemize}

\noindent Specific to the policy gradient approaches:
\begin{itemize}
    \item Samples between updates: $N=200$
    \item Policy training batch size: all samples
\end{itemize}

\noindent Specific to the value based approaches:
\begin{itemize}
    \item Samples between updates $N=20$
    \item Replay buffer size: \num{64000}
\end{itemize}




\end{document}